\documentclass{article}

\usepackage{microtype}
\usepackage{graphicx}
\usepackage{subfigure}
\usepackage{booktabs} 
\usepackage{ltablex}
\usepackage{url}            
\usepackage{booktabs}       
\usepackage{amsfonts}       
\usepackage{nicefrac}       
\usepackage{microtype}      
\usepackage[normalem]{ulem}
\usepackage[colorinlistoftodos]{todonotes}
\usepackage{mathrsfs}
\usepackage{amssymb}
\usepackage{dirtytalk}
\usepackage{hyperref}

\usepackage[arxiv]{icml2019}

\newcommand{\untrain}{AAL}

\newcommand{\umaml}{AAL-MAML++}
\newcommand{\maml}{MAML++}
\newcommand{\proto}{ProtoNets}
\newcommand{\uproto}{AAL-ProtoNets}
\makeatletter

\newcommand{\mypm}{\mathbin{\mathpalette\@mypm\relax}}
\newcommand{\@mypm}[2]{\ooalign{%
  \raisebox{.1\height}{$#1+$}\cr
  \smash{\raisebox{-.6\height}{$#1-$}}\cr}}
\makeatother

\icmltitlerunning{Assume, Augment and Learn: Unsupervised Few-Shot Meta-Learning via Random Labels and Data Augmentation (AAL)}

\begin{document}

\twocolumn[
\icmltitle{Assume, Augment and Learn: Unsupervised Few-Shot Meta-Learning via Random Labels and Data Augmentation}



\icmlsetsymbol{equal}{*}

\begin{icmlauthorlist}
\icmlauthor{Antreas Antoniou}{ed}
\icmlauthor{Amos Storkey}{ed}
\end{icmlauthorlist}

\icmlaffiliation{ed}{School of Informatics, University of Edinburgh, Edinburgh, UK}

\icmlcorrespondingauthor{Antreas Antoniou}{a.antoniou@sms.ed.ac.uk}

\icmlkeywords{Machine Learning, ICML}

\vskip 0.3in
]



\printAffiliationsAndNotice{}  

\begin{abstract}
The field of few-shot learning has been laboriously explored in the supervised setting, where per-class labels are available. On the other hand, the unsupervised few-shot learning setting, where no labels of any kind are required, has seen little investigation. We propose a method, named \emph{Assume, Augment and Learn} or \emph{\untrain}, for generating few-shot tasks using unlabeled data. We randomly label a random subset of images from an unlabeled dataset to generate a support set. Then by applying data augmentation on the support set's images, and reusing the support set's labels, we obtain a target set. The resulting few-shot tasks can be used to train any standard meta-learning framework. Once trained, such a model, can be directly applied on small real-labeled datasets without any changes or fine-tuning required. In our experiments, the learned models achieve good generalization performance in a variety of established few-shot learning tasks on Omniglot and Mini-Imagenet.
\end{abstract}

\section{Introduction}
Much of the success of modern deep learning relies on supervised learning \citep{krizhevsky2012imagenet,he2016deep,huang2017densely,ravi2016optimization,santoro2016one,van2016wavenet}. Supervised learning describes a family of learning algorithms that, given a set of input-output pairs, can learn a model that can map a set of input data to a set of output labels. A good supervised model should not only be able to predict the correct output for the data it was trained on (i.e. the training set) but also have high generalization power (i.e. to be able to predict the correct outputs for previously unseen data points). 

\begin{figure}[ht]
    \centering
    \includegraphics[width=0.99\columnwidth]{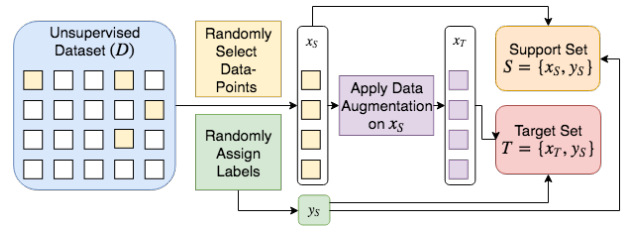}
    \caption{Proposed method.  We follow the set-to-set \cite{vinyals2016matching} few-shot learning framework, where during inference time, a learner acquires task-specific information from a \textbf{support set}, and is then evaluated on a \textbf{target set} for that task. Building support and target sets for few-shot tasks ordinarily requires labelled information. However, to produce such tasks in the absence of labelled data, we randomly select some data points from a label-free dataset and assign random labels to them, in order to form the support set. Then, the support set data-points are augmented, and the label information kept the same to form the target set.} 
    \label{fig:unsupervised_task_sampling}
\vspace{-7.0mm}
\end{figure}

In general, to train a deep neural network model via supervised learning, one needs to have a large number of input-output pairs available. Labels are usually acquired by human annotators, and the labelling process is laborious. There have been attempts at semi-automating the process \cite{parkhi2015deep} but even those require a strong pretrained model to be available, which needs substantial data in order to be trained in the first place. The data labeling component of data acquisition is considered by many to be the most costly and time consuming part of the machine learning pipeline. In stark contrast, for many problems, unlabeled data is freely available in large quantities in the internet. 

In the context of few-shot learning, the fact that one can train a robust few-shot learner using just 24K handwritten digit samples (i.e. the Omniglot training dataset) that can achieve over 95\%+ on a variety of tasks is impressive \cite{finn2017model,snell2017prototypical,ravi2016optimization,antoniou2018train}. However, a well designed machine learning algorithm should be able to utilize other larger unlabeled datasets, related to \textit{but not identical to} the test scenario, by using them to meta-learn robust few-shot learning models. A robust unsupervised few-shot learner, once trained, should be able to take new and different data from a task with labels, acquire task-specific knowledge from the test-time support set and generalize well on the test-time target set.



In this paper we propose such a method for leveraging unsupervised data to generate tasks for few-shot learners. In summary, we create support sets by giving random labels to subsets of the unsupervised data and generate corresponding target sets through data-augmentation transformations of the support set data points. Our unsupervised meta-learning must then learn networks that transfer consistent to data augmented examples. This can then be applied to real labelled data in the test setting (see Figure~\ref{fig:unsupervised_task_sampling}).

The approach is evaluated on two separate types of few-shot learning models, the Model Agnostic Meta Learning framework \cite{finn2017model} and the Prototypical Networks framework \cite{snell2017prototypical}. Once a few-shot learner is trained using the tasks generated with the proposed method, it can then be used with real-labeled (few-shot) data directly, without any fine tuning, and generalize very well on the target task. More specifically, the technique involves building a support set and target set using only unlabeled data. To build a support set, we sample a number of samples from our training dataset and then, since no labels are available, we \emph{assume} labels by randomly generating labels for our support set images. In the supervised setting the target set contains new (i.e. that do not exist in the support set) instances of the same classes found in the support set. To generate such a target set, we apply data augmentation methods to our support set, hence creating a target set with semantically the same classes as our support set, but with a different enough perspective on those samples to be a good generalization evaluation set.

The motivation for our method for unsupervised few-shot learning comes from the human ability of finding patterns among clusters of data items, even when the data items are being repeatedly re-arranged into new clusters. Humans are very good at finding patterns everywhere thus no matter how one chooses to cluster data items, an average healthy human being should be able to find features that can effectively cluster those data items correctly as shown in figure \ref{fig:random_clusters}. Following that idea we attempt to train models by assuming a clustering for a given support set, then generating a target set by augmenting the support set, and training our model using the MAML framework such that the model can acquire fast knowledge on the support set and generalize well on the target set.



In the following sections, we first describe our setting, then the MAML framework on which we built our method, before proceeding to the description of our method and the evaluation experiment types used. Finally we present our empirical results and draw our conclusions.

\section{Related Work}

The reformulation of few-shot learning as a meta-learning problem using the \emph{set-to-set} \cite{vinyals2016matching} few-shot learning setting was arguably one of the main enablers of the substantial progress that has been observed in the field of few-shot learning in the past few years. 


To cast few-shot learning as a meta-learning problem, a batch of \emph{samples} is replaced with a batch of \emph{tasks}. Each task contains a dataset, which is further divided into two separate sets, a \emph{support set} used by the model for acquiring task-specific information and a \emph{target-set} used to evaluate the generalization performance of the model on that particular task. More specifically, a support set is composed of a pre-specified number of samples ($K$) from a number of classes ($N$), and a target set is composed from different (i.e. not found in the support set) instances of the same classes as found in the support set. Both sets contain label information for their respective data-points. The tasks used at test time are composed of (sample) classes that were not used during training. This is done to evaluate the model's generalization performance on previously unseen tasks. Meta-learning is about learning \emph{how} to learn such that a system can transfer well to previously unseen settings, thus testing on previously unseen tasks is of prime importance if one is to acquire a good measure of a meta-learning model's performance. The authors also proposed a short notation for describing set-to-set tasks. Each task set has a number of classes indicated by $N$, and a number of samples per class indicated by $K$, so one can instead refer to this setting as the \emph{N-way, K-shot} setting.

Once the set-to-set few-shot learning setting was introduced, a variety of meta-learning systems utilizing it were introduced. The first of which was Matching Networks \cite{vinyals2016matching} where both the support and target sets are embedded down to a low-dimensional space using two learnable \emph{embedding functions} $g$ (used on the support set) and $f$ (used on the target set) parameterized as neural networks. Once the low-dimensional embeddings are computed, the target set items are compared with the support set items using a non-parametric distance metric such as cosine distance. Once the distance vectors are acquired, one can apply a softmax function over them, to acquire a probability distribution that expresses the class of a particular target set item over the support set classes.  

After embedding-based meta-learning models came the resurgence of gradient-based meta-learning models. Such models utilize inference-time model-state updates to acquire task-specific knowledge from a support set, such that they can generalize strongly on a particular task's target set. The first of such approaches was the \emph{meta-learner} LSTM \cite{ravi2016optimization} which jointly learns a gradient conditional weight update function and a parameter initialization for a \emph{base} model. At inference time, the learned weight update function is used to apply a single update step on the base-model using gradients with respect to a support set loss. Then, the updated base-model is applied on a target set to compute a target loss, which is then used as the system's optimization loss. 

Subsequently to Meta-Learner LSTM the \emph{Model Agnostic Meta Learning} \cite{finn2017model} (MAML) framework was introduced. In MAML the authors propose to use standard stochastic gradient descent instead of a learnable update function and, in addition, to increase the number of update steps the model is allowed to take on a task's support set. By doing so, MAML achieved state-of-the-art performance in the supervised few-shot learning setting across all established few-shot tasks in both Omniglot and Mini-Imagenet. However, MAML also exhibited a significant amount of problems, including training instability problems, significant sensitivity to architecture selection and requirement for extensive hyperparameter tuning for it to achieve state-of-the-art results. Furthermore, many details in the design of MAML were chosen without much consideration for the multi-step nature of the model, which constrained the model's optimization process unnecessarily and in a way that reduced it's potential generalization performance and convergence speed. The authors in \citet{antoniou2018train} propose various modifications for MAML that resolve its instability problems, enable automatic learning of it's hyperparameters for a given task and unshackle MAML from its design-derived contraints. The resulting model called MAML++ achieves a significant improvement in the generalization performance across all established few-shot tasks whilst decreasing the model's computational overheads. 

Unsupervised deep learning has been extensively investigated in the context of generative models \cite{doersch2016tutorial, vincent2008extracting,goodfellow2014generative}, yet, has only been briefly attempted in classification models    \cite{caron2018deep,bojanowski2017unsupervised}. 

Concurrent work in unsupervised few-shot learning was introduced in the form of CACTUs \cite{hsu2018unsupervised}, where the authors propose the generation of support and target sets using k-means clustering in combination with semantically meaningful representations. Furthermore, work has also been carried out in the semi-supervised setting, which attempts to learn an unsupervised loss function \cite{semifewmaml2018}, such that a MAML-based model can learn from an unlabelled support task and generalize well on a target task. Furthermore, days prior to this paper's submission we became aware of concurrent work carried out in UMTRA \cite{unfewimagevideo}, where the authors use an approach very similar to ours to meta-learn few-shot models that generalize well to real-labelled datasets. Our work and theirs differs in that we conducted extensive ablation studies on data-augmentation strategies, ranging from very simple crops all the way to warping \cite{wong2016understanding} and cutout \cite{devries2017improved}. Furthermore, we evaluate our method on the \maml\ model instead of the original MAML model. In terms of results, our results outperform UMTRA on the Omniglot tasks, whereas UMTRA outperforms our method by utilizing the data-augmentations learned in the AutoAugment \cite{cubuk2018autoaugment} work. However, the augmentations in AutoAugment were learned over a large subset of ImageNet, hence, utilizing AutoAugment might be considered as transfering knowledge from a model trained on a vastly larger labelled dataset. Which, in a way, defeats the purpose of unsupervised learning. 

\section{Setting}\label{unmaml-setting}
The setting used is one where, given an unsupervised dataset we want to learn a few-shot learning model that can generalize very well on small few-shot labeled dataset-tasks.
Thus we make use of three sets. The first set is the \emph{meta-training} set, which is label-free and used to train a few-shot learner using the proposed unsupervised method. The second and third sets, are called the \emph{meta-validation} and \emph{meta-test} sets respectively, and contain labeled data, to be used to create evaluation few-shot tasks. Using a validation set to pick the best train model and a test set to produce the final test errors removes any potential unintended (human-derived/implicit) over-fitting. Intuitively, the meta-validation and meta-test sets are generated in the exact same manner that the validation and test tasks are generated in any other few-shot learning methodologies such as MAML \cite{finn2017model} and Prototypical Networks \cite{snell2017prototypical}.

Training meta-learning models requires using a large number of \emph{tasks}. To generate a few-shot task we use the \emph{set-to-set} few-shot learning scheme proposed in \citep{vinyals2016matching}. In the set-to-set few-shot learning setting, a task is composed by a training or \emph{support} set and a validation or \emph{target} set. The support set consists of a number of classes ($N$), and a small number of data-samples per class ($K$). The target set consists from samples from the same classes as the support set, but using different instances of those classes than the ones in the support set. A setting of $N$ classes per set and $K$ samples per class can alternatively be refereed using the shortened notation of \emph{N-way, K-shot learning} setting.

\section{Motivation}
Initial motivation for the work came from the human ability of being able to find features that can accurately describe a set of randomly clustered data-points, even when the clusters are continuously randomly reset. Transferring this setting to a machine learning setting could produce strong representations for classification.

\begin{figure}[ht]
	\centering
	\includegraphics[width=0.73\columnwidth]{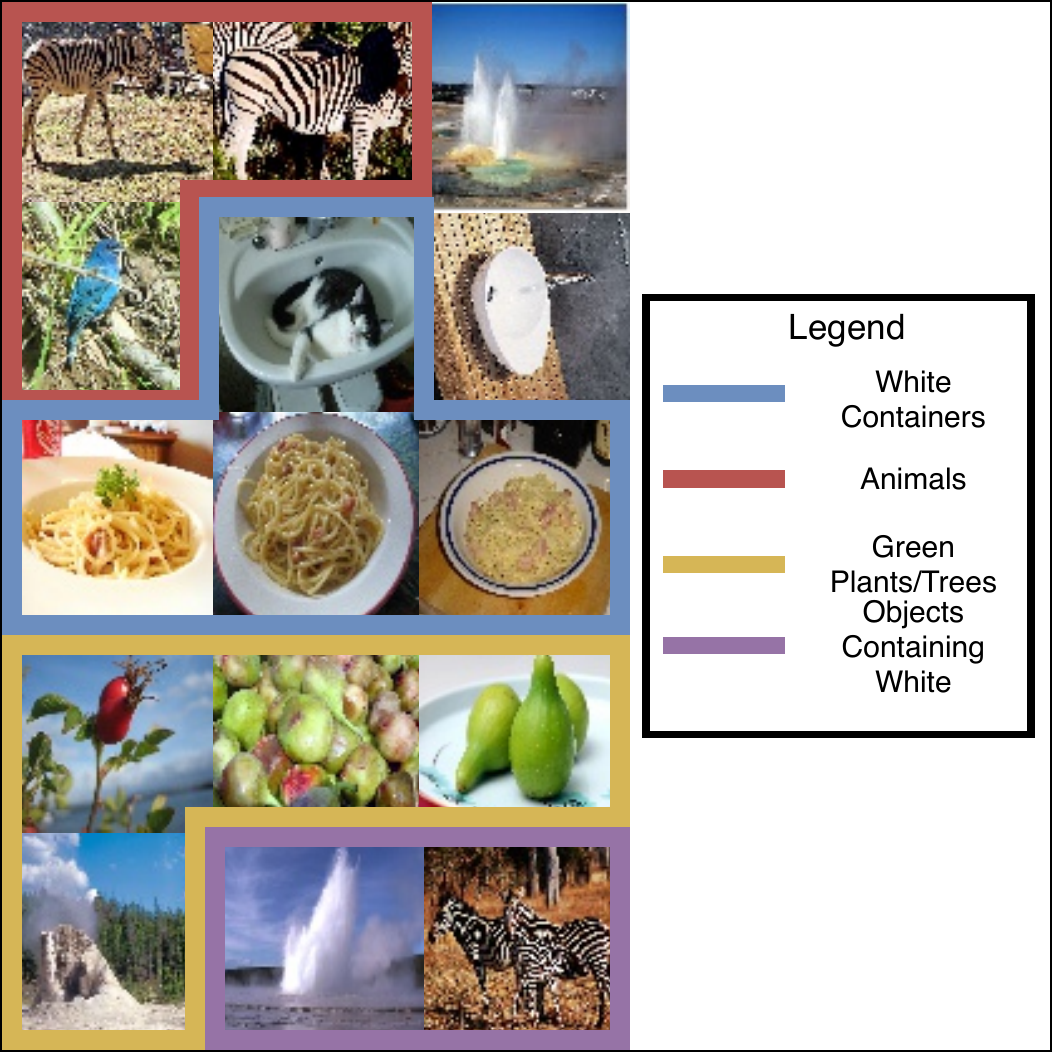}
	\caption{Random Clusterings and Features: In this figure, one can see the concept of random labels and how useful features can be learned by random labels alone. For example, the items contained within the blue "Tee" shape can be assigned the labeling of white containers, thus learning features for white containers would allow a classifier to learn sensible parameters.}
	\label{fig:random_clusters}
\vspace{-8mm}
\end{figure}

However, using random clusters in a standard (base) deep-learning setting would not work very well since the model is trying to learn a mapping from $x$ to $y$, that holds for all seen $x$ to $y$ pairs. So if one were to randomly reset the labels ($y$) as the training progresses to create new tasks, the model would not be able to arrive to a solution that will generalize across all $x$ to $y$ pairs since many of them will be contradictory. However, intuitively, it should work in in a meta-learning setting. In a meta-learning setting the goal is to be able to learn a model that given a \emph{task}, can at inference time, acquire task-specific knowledge from a support set such that it can do well in that one task's target set, before throwing away the fast-knowledge acquired. So, intuitively, one could think that learning a meta-model that captures across-task similarities but can throw away task-specific information (such as the specific class clusters for a given task) would work very well. This is what this paper is about. Exploiting semantic similarities between data points to learn robust across-task representations that can then be used in a setting where supervised labels are available. Figure \ref{fig:random_clusters} illustrates the concept of how random clusters can enable the learning of sensible features.

\section{Generating Tasks for Unsupervised Few-Shot Learning}\label{unmaml}
To enable unsupervised learning in a few-shot classification setting, one must find a way to generate tasks. As defined in \ref{unmaml-setting}, a task is composed of a support and target set. A support set consists of samples from a number of distinct classes and their associated labels. Since no label information is available, we instead \emph{assume} labels by randomly assigning labels for a randomly sampled set of data-points. This will effectively cluster the data items into groups. Since the model takes update steps on the support set, it should be able to acquire fast-knowledge in its parameters that classify that support set very well. Generating a semantically related (to the support set) target set is where much of the difficulty lies. A target set should have \emph{different} instances of the \emph{same} classes appearing in the support set. Since we randomly assigned classes to the data-points we have no information on the semantics connecting the samples together in a class. To overcome this problem we generate a target set by \emph{augmenting} the support set data-points using data augmentation. Doing so sets the classes in the support set to perfectly match to the classes in the target set, but the samples in the two sets are different enough to allow the target set to serve as a good evaluation set. In algorithms \ref{alg:supervised-sampling} and \ref{alg:unsupervised-sampling} one can see how a supervised and unsupervised sampling procedure works.

\begin{algorithm}[]
	\caption{Supervised MAML Sampling Strategy}\label{alg:supervised-sampling}
	\begin{algorithmic}[1]
    	\STATE \textbf{Require:} Dataset $\mathcal{D}$ with $\mathcal{C}$ number of classes and $\mathcal{M}$ samples per class
		\STATE Sample $N$ class indexes from $\mathcal{D}$, where $N \leqslant \mathcal{C}$
        \STATE Sample the support set $S$ by sampling $K$ samples from each selected class where $K \leqslant \mathcal{M}$
        \STATE Sample the target (evaluation) set $E$ by sampling $J$ samples from each selected class, making sure that these samples do not contain any support set images  where $E \leqslant \mathcal{M}$ and ${E} \cap {K} = \emptyset$
        \STATE \textbf{Return} $S$, $E$
	\end{algorithmic}
\end{algorithm}

\begin{algorithm}[]
	\caption{Unsupervised MAML Sampling Strategy}\label{alg:unsupervised-sampling}
	\begin{algorithmic}[1]
    	\STATE \textbf{Require:} Dataset $\mathcal{D}$ with $I$ number of data-points
		\STATE Sample $N \times K$ data-points from $\mathcal{D}$, where $N$ is the number of classes per set\footnotemark and $K$ is the number of samples per class $(N \times K) \leqslant I$
        \STATE Build the support set $S$ by assigning random labels to the previously $N \times K$ sampled data-points
        \STATE Build the target (evaluation) set $E$ by augmenting the support set $S$ samples and keeping the labels identical
        \STATE \textbf{Return} $S$, $E$
	\end{algorithmic}
\end{algorithm}

 \footnotetext{One should choose the same number of classes and number of samples per class as for the supervised setting the model will be tested on at inference time.}

\section{Meta-Learning Frameworks used for Evaluation}
\untrain\ can be applied on any existing few-shot meta-learning technique, as long as the method can be trained using the set-to-set few-shot learning framework. To demonstrate this, we apply our method on two very popular and effective few-shot meta-learning frameworks, the Prototypical Networks framework and the MAML framework. 

\emph{Prototypical Networks}:
Prototypical Networks \cite{snell2017prototypical} were introduced in an attempt to improve performance and reduce computational and storage requirements of Matching Networks. In Prototypical Networks, the authors propose using per-class or \emph{prototype} embeddings instead of per-sample embeddings. More specifically they propose computing the mean of per-class embeddings, hence obtaining a single embedding vector that summarizes a particular class. Once the prototypical embeddings are computed, they are compared with a target sample's embedding, using a suitable distance metric, such as euclidean or cosine distance, to obtain a probability distribution over the prototypical classes predicting the target sample's class membership. Not only does the usage of prototypical embeddings improve the generalization performance of the system, but it also reduces both the computational and space complexity of the system. In particular, using prototypical embeddings renders the computational complexity of computing the similarity vectors linearly proportional to number of classes ($N$) instead of number of samples ($N \times K$), hence reducing the computational complexity in all cases, except the case when $K = 1$. Furthermore, the storage complexity is also reduced from $N \times K$ to $N$. 

\emph{Model Agnostic Meta-Learning} \cite{finn2017model} or (MAML) is a meta-learning framework for few-shot learning. It approaches few-shot learning by learning an initialization for the parameters of a base-model. Given a task, MAML computes the loss and gradients of a base-model with respect to the support set and takes a step in the direction of the gradients, thus acquiring task-specific knowledge usually refered to as the \emph{inner-loop} optimization process. This process is repeated a number of times. Once the inner loop update steps have been completed, the base model is evaluated on the task's target set to obtain the target set loss. Once computed, the target loss, is used to update the base-model's initialization, by taking a step in the direction that minimizes the target loss. Effectively, this trains a base-model parameter initialization, that after a number of SGD updates wrt a support set, the model can generalize very well on a target set. 

\subsection{Selecting Target Set Data Augmentations}\label{label-free-data-augmentation}

Selecting the right combination of data augmentation techniques is crucial for our proposed approach. Selecting a data augmentation method which makes minimal changes to the data-points could produce samples that are too similar to the originals (thus being "too easy") for the model and thus causing reduced generalization performance. On the other hand using a data augmentation that changes the data-points too much could change semantics within the image that are important for the class of the image. Removing semantically important features from the image can make learning very hard, and in addition, guide the network to learn a function that does not generalize well to real-labeled tasks. 

\begin{table*}[!htp]
\begin{tabular}{lll}
\hline
\multicolumn{1}{|l|}{Name}                        & \multicolumn{1}{l|}{Omniglot Hyperparameters}                 & \multicolumn{1}{l|}{Mini-Imagenet Hyperparameters}                   \\ \hline
\multicolumn{1}{|l|}{Random Crops (C)}            & \multicolumn{1}{l|}{Padding: 7 pixels}                        & \multicolumn{1}{l|}{Padding: 21 pixels}                         \\ \hline
\multicolumn{1}{|l|}{Random Horizontal Flips (H)} & \multicolumn{1}{l|}{Flip probability 50\%}                     & \multicolumn{1}{l|}{Flip probability 50\%}                       \\ \hline
\multicolumn{1}{|l|}{Random Vertical Flips (V)}   & \multicolumn{1}{l|}{Flip probability 50\%}                     & \multicolumn{1}{l|}{Flip probability 50\%}                       \\ \hline
\multicolumn{1}{|l|}{Random Rotations (R)}        & \multicolumn{1}{l|}{1 - 30 degrees}                           & \multicolumn{1}{l|}{1 - 270 degrees}                            \\ \hline
\multicolumn{1}{|l|}{Image-Pixel Dropout (DROP)}  & \multicolumn{1}{l|}{Drop probability: 30\%}                    & \multicolumn{1}{l|}{Drop probability: 70\%}                      \\ \hline
\multicolumn{1}{|l|}{Image-Pixel Cutout (CUT)}    & \multicolumn{1}{l|}{Holes: 5, Side Length: 4 - 14 pixels}     & \multicolumn{1}{l|}{Holes: 5, Side Length: 11 - 42 pixels}      \\ \hline
\multicolumn{1}{|l|}{Image-Warping (W)}           & \multicolumn{1}{l|}{Warping area: 14 + U(0, 6) x 14 + U(0,6)} & \multicolumn{1}{l|}{Warping area: 42 + U(0, 41) x 42 + U(0,41)} \\ \hline
\multicolumn{1}{|l|}{Random Grayscale (G)}        & \multicolumn{1}{l|}{Not used}                                 & \multicolumn{1}{l|}{Grayscale probability: 50\%}                 \\ \hline
                                                                                                                 
\end{tabular}
\caption{This table presents the types and settings of data augmentation used to generate target sets in our experiments.}
\label{table:data-augmentations}
\end{table*}

Table \ref{table:data-augmentations}, presents the list of data augmentations that we used to generate our target sets in our experiments. In more detail we employed random crops, random rotations, random horizontal and vertical flips, randomly zeroing out a fraction of the image pixels or \emph{Image-Pixel Dropout} \cite{krizhevsky2012imagenet}, cutout, warping and randomly gray-scaling our image. Image-pixel dropout proved to be potentially the most important augmentation, since it produced the most generalizable and transferable models. 

Due to limitations in compute, we decided that we would only search through combinations of rotations, cutout, dropout, warping and grayscaling of images, whilst keeping random crops, horizontal and vertical flips shared across all experiments. Doing so exponentially reduced the number of required experiments. Preliminary experiments with crops, horizontal and vertical flips demonstrated that their combination improved the generalization performance of models across all few-shot tasks across both Omniglot and Mini-Imagenet, hence their usage as shared (and baseline) augmentations.

\begin{table*}[!htb]
\begin{center}
\begin{tabular}{|l|l|l|l|l|}
\hline
                              & \multicolumn{2}{l|}{5-way Accuracy (val/test)}        & \multicolumn{2}{l|}{20-way Accuracy(val/test) } \\ \hline
Omniglot                                  & 1-shot             & 5-shot            & 1-shot             & 5-shot                   \\ \hline
Randomly initialized                      & $70.3\mypm0.36\%$  &$86.9\mypm0.27\%$  & $44.7\mypm0.45\%$  &$62.1\mypm0.85$\\\hline
\umaml + CHV                              & $75.8\mypm0.63\%$  &$96.6\mypm0.32\%$  & $42.0\mypm1.16\%$  &$76.6\mypm1.14$\\\hline
\umaml + CHVW                             & $88.4\mypm0.75\%$  &$98.0\mypm0.32\%$  & $70.2\mypm0.86\%$  &$88.3\mypm1.22$\\\hline
\umaml + CHVR                             & $78.0\mypm2.39\%$  &$96.6\mypm0.04\%$  & $58.9\mypm1.16\%$  &$81.6\mypm0.82$\\\hline
\umaml + CHV + DROP                       & $85.7\mypm1.09\%$  &$97.1\mypm0.12\%$  & $65.1\mypm0.62\%$  &$83.8\mypm0.62$\\\hline
\umaml + CHV + CUT                        & $74.0\mypm1.60\%$  &$95.9\mypm0.44\%$  & $54.0\mypm1.59\%$  &$75.2\mypm0.62$\\\hline
\umaml + CHV + DROP + CUT                 & $85.7\mypm1.09\%$  &$95.7\mypm0.38\%$  & $68.3\mypm0.63\%$  &$85.8\mypm0.62$\\\hline
\umaml + CHVR + DROP                      & $86.16\mypm1.09\%$ &$96.5\mypm0.87\%$  & $64.6\mypm0.39\%$  &$82.0\mypm0.19$\\\hline
\umaml + CHVR + CUT                       & $77.8\mypm1.00\%$  &$95.1\mypm1.05\%$  & $60.2\mypm0.66\%$  &$80.6\mypm0.53$\\\hline
\umaml + CHVR + CUT + DROP                & $85.6\mypm1.05\%$  &$96.7\mypm0.45\%$  & $66.5\mypm0.27\%$  &$83.1\mypm0.64$\\\hline
\maml (supervised upper bound)            & $99.4\mypm1.05\%$  &$99.9\mypm0.45\%$  & $97.7\mypm0.27\%$  &$99.1\mypm0.64$\\\hline

\end{tabular}
\end{center}
\vspace{-4mm}
\caption{\umaml\ Omniglot Ablation studies}
\label{table:umaml-ablation-omniglot}
\end{table*}

\begin{table*}[!ht]
\begin{center}
\begin{tabular}{|l|l|l|l|l|}
\hline
                              & \multicolumn{2}{l|}{5-way Accuracy (val/test)}        & \multicolumn{2}{l|}{20-way Accuracy(val/test) } \\ \hline
Omniglot                                         & 1-shot             & 5-shot            & 1-shot              & 5-shot            \\ \hline
Randomly initialized                             & $25.72\mypm0.22\%$& $28.03\mypm0.28\%$ & $07.31\mypm0.41\%$& $08.04\mypm0.36\%$  \\ \hline
\uproto + CHV                              & $84.23\mypm0.93\%$ & $89.14\mypm0.36\%$& $64.34\mypm1.07\%$& $74.28\mypm1.76\%$  \\ \hline
\uproto + CHV  + CUT                        & $82.78\mypm2.25\%$ & $89.04\mypm0.09\%$& $64.23\mypm1.06\%$& $71.37\mypm1.18\%$  \\ \hline
\uproto + CHV  + CUT + DROP                 & $83.38\mypm0.60\%$ & $88.98\mypm0.88\%$& $66.26\mypm0.96\%$& $73.94\mypm1.18\%$  \\ \hline
\uproto + CHVR + CUT                    & $82.06\mypm0.53\%$ & $87.78\mypm0.21\%$& $64.77\mypm1.46\%$& $70.66\mypm0.53\%$  \\ \hline
\uproto + CHV  + DROP                       & $84.66\mypm0.70\%$ & $88.41\mypm0.27\%$& $68.79\mypm1.03\%$& $74.05\mypm0.46\%$  \\ \hline
\uproto + CHVR + DROP + CUT             & $81.24\mypm0.62\%$ & $87.15\mypm0.86\%$& $65.31\mypm0.76\%$& $72.04\mypm0.33\%$  \\ \hline
\uproto + CHVR + DROP                   & $82.65\mypm0.36\%$ & $88.13\mypm0.75\%$& $67.10\mypm1.13\%$& $71.92\mypm0.41\%$  \\ \hline
\uproto + CHVR                          & $82.12\mypm0.97\%$ & $86.04\mypm1.09\%$& $64.39\mypm1.27\%$& $70.39\mypm0.99\%$  \\ \hline
\proto (supervised upper bound)         & $98.92\mypm0.07\%$ & $99.56\mypm0.07\%$& $97.87\mypm0.11\%$& $98.90\mypm0.10\%$  \\ \hline

\end{tabular}
\caption{\uproto\ Omniglot ablation studies.}
\label{table:uproto-ablation-omniglot}
\end{center}
\vspace{-7mm}
\end{table*}


\begin{table*}[!ht]
\begin{center}
\begin{tabular}{|l|l|l|l|l|}
\hline
                        & \multicolumn{2}{l|}{5-way Accuracy (test)}  \\ \hline
Mini-Imagenet                                     & 1-shot             &5-shot           \\ \hline  
Randomly initialized                              &$28.55\mypm0.22\%$  &$36.68\mypm0.23\%$ \\ \hline  
\umaml + CHV                                &$33.06\mypm0.49\%$  &$40.75\mypm1.03\%$\\\hline
\umaml + CHVR                     &$33.21\mypm0.90\%$  &$40.34\mypm1.64\%$\\\hline
\umaml + CHV  + CUT                       &$33.34\mypm0.54\%$  &$39.44\mypm0.94\%$\\\hline
\umaml + CHV  + DROP                      &$30.86\mypm0.64\%$  &$40.41\mypm0.68\%$\\\hline
\umaml + CHVW                       &$33.30\mypm0.31\%$  &$46.98\mypm0.16\%$\\\hline
\umaml + CHVWG                      &$34.57\mypm0.74\%$  &$49.18\mypm0.47\%$\\\hline
\umaml + CHVR + CUT            &$33.09\mypm0.51\%$  &$40.11\mypm0.76\%$\\\hline
\umaml + CHVR + DROP           &$31.70\mypm0.54\%$  &$39.38\mypm0.57\%$\\\hline
\umaml + CHV  + DROP + CUT             &$31.55\mypm0.85\%$  &$38.76\mypm0.83\%$\\\hline
\umaml + CHVR + DROP + CUT &$31.44\mypm0.78\%$  &$39.87\mypm0.89\%$\\\hline
\maml (supervised upper bound)                    &$52.15\mypm0.26\%$  &$67.87\mypm0.42\%$\\\hline

\end{tabular}
\caption{\umaml ablation studies.}
\label{table:umaml-ablation-mini-imagenet}
\end{center}
\vspace{-4mm}
\end{table*}

\begin{table*}[]
\begin{center}
\begin{tabular}{|l|l|l|l|l|}
\hline
                        & \multicolumn{2}{l|}{5-way Accuracy (test)}  \\ \hline
Mini-Imagenet                & 1-shot           & 5-shot           \\ \hline
Randomly initialized                     & $22.96\mypm0.05\%$ & $24.73\mypm0.10\%$      \\\hline
\uproto + CHV                      & $37.67\mypm0.39\%$ & $40.29\mypm0.68\% $\\\hline
\uproto + CHV + CUT                & $36.38\mypm0.91\%$ & $40.89\mypm1.30\%$ \\\hline
\uproto + CHV + CUT + DROP         & $33.13\mypm0.95\%$ & $36.64\mypm0.79\%$ \\\hline
\uproto + CHVR + CUT + DROP     & $31.93\mypm0.28\%$ & $36.45\mypm1.33\%$ \\\hline
\uproto + CHVR + CUT            & $33.92\mypm0.60\%$ & $39.87\mypm0.53\%$ \\\hline
\uproto + CHV + DROP               & $32.12\mypm0.34\%$ & $36.12\mypm1.11\%$ \\\hline
\uproto + CHVR + DROP           & $31.13\mypm0.48\%$ & $36.83\mypm0.57\%$ \\\hline
\uproto + CHVR                  & $34.28\mypm0.21\%$ & $39.83\mypm0.85\%$ \\\hline
\proto Protonet (supervised upper bound) & $50.16\mypm0.15\%$ & $65.56\mypm0.13\%$ \\\hline

\end{tabular}
\caption{\uproto\ ablation studies.}
\label{table:uproto-ablation-mini-imagenet}
\end{center}
\vspace{-5mm}
\end{table*}

\begin{table*}[!ht]
\begin{center}
\begin{tabular}{@{}lcccc@{}}
\toprule
Algorithm \hfill (way, shot) & (5, 1) & (5, 5) & (20, 1) & (20, 5) \\
\midrule
Training from scratch & 52.50\% & 74.78\% & 24.91\% & 47.62\% \\
ACAI CACTUs-MAML       & 68.84\% & 87.78\% & 48.09\% & 73.36\% \\
ACAI CACTUs-ProtoNets  & 68.12\% & 83.58\% & 47.75\% & 66.27\% \\
BiGAN CACTUs-MAML       & 58.18\% & 78.66\% & 35.56\% & 58.62\% \\
BiGAN CACTUs-ProtoNets  & 54.74\% & 71.69\% & 33.40\% & 50.62\% \\
UMTRA                   & 77.80\% & 92.74\% & 62.20\% & 77.50\% \\
\uproto\ (ours)              & 84.66\% & 89.14\% & 68.79\% & 74.28\% \\
\umaml\ (ours)              & \textbf{88.40\%} & \textbf{97.96\%} & \textbf{70.21\%} & \textbf{88.32\%} \\
\midrule
MAML (supervised upper bound) & 94.46\% & 98.83\% & 84.60\% & 96.29\% \\
\maml (supervised upper bound)\ & 99.42\% & 99.93\% & 97.76\% & 99.12\% \\
\proto\ (supervised upper bound) & 98.92\% & 99.56\% & 97.87\% & 98.90\% \\
\bottomrule
\end{tabular}
\caption{Comparative Omniglot results.}
\label{table:compare-omniglot}
\end{center}
\vspace{-8mm}
\end{table*}

\begin{table}
\begin{center}
\begin{tabular}{@{}lcc@{}}
\toprule
Algorithm \hfill (way, shot) & (5, 1) & (5, 5) \\ 
\midrule
Training from scratch & 27.59\% & 38.48\% \\
DeepCluster CACTUs-MAML & \textbf{39.90\%} & \textbf{53.97\%} \\
DeepCluster CACTUs-ProtoNets & 39.18\% & 53.36\% \\
UMTRA + AutoAugment          & \textbf{39.93\%} & 50.73\% \\
\uproto                      & 37.67\% & 40.29\% \\
\umaml                       & 33.30\% & 49.18\% \\
\midrule
MAML (supervised upper bound) & 46.81\% & 64.13\% \\
\maml (supervised upper bound) & 52.15\% & 67.87\% \\
ProtoNets (supervised upper bound) & 50.16\% & 65.56\% \\
\bottomrule
\end{tabular}
\caption{Comparative Mini-Imagenet results.}
\label{table:compare-mini-imagenet}
\end{center}
\vspace{-10mm}
\end{table}


\section{Datasets}\label{dataset-umaml}
To evaluate our methods we used the Omniglot and Mini-Imagenet datasets. As discussed in Section \ref{unmaml-setting} we split our dataset in a background (or meta-training) set and an evaluation set (which is then further split into meta-validation and meta-test sets). The Omniglot dataset is composed of 1623 character classes from various alphabets. There exist 20 instances of each class in the dataset. For meta-training, meta-validation and meta-test sets we use classes 1-1150, 1150-1200 and 1200-1623 respectively.
The Mini-Imagenet dataset was proposed in \citet{ravi2016optimization}, it consists of 600 instances of 100 classes from the ImageNet dataset, scaled down to 84x84. We use the split proposed by the authors, which consists of 64 classes for training, 12 classes for validation and 24 classes for testing. We use the training, validation and testing sets as the meta-training, meta-validation and meta-test respectively for our experiments.

\section{Experiments}

To evaluate the generalization performance of the unsupervised trained models on supervised few-shot tasks we ran experiments where we trained our model using the \untrain\ method and after each unsupervised meta-training epoch we applied the resulting model without any additional fine tuning on the supervised meta-validation set. Once all epochs were completed, the best validation model was then applied on the test set which produced the final test performance of the model.

We applied \untrain\ on two separate model types. The first, was the MAML++ \cite{antoniou2018train} model (an improved variant of MAML), more specifically, we used the LSLR (Per-Layer Per-Step Learning Rates), along with the per-step batch norm modifications (BNWB/BNRS) and the multi-step loss optimization (MSL) to both stabilize training and improve the generalization performance. 
The second, was the prototypical networks \cite{snell2017prototypical} model. 

To apply \untrain\ on a given model, we simply sampled training tasks using the \untrain\ sampling scheme, and then evaluated our model on validation and test tasks using the supervised meta-learning formulation as done in \citet{finn2017model}. As a result, the evaluation metrics would effectively be quantifying the transferability of the representations learned using unsupervised data, on a real-labeled supervised meta-learning task, rendering our evaluation results directly comparable to supervised meta-learning models. 

\subsection{Neural Network Architectures}
The architecture used for the base classifier was the one proposed in \citet{finn2017model}. A convolutional neural network with 4 layers. Each layer is composed of a convolutional layer followed by batch normalization followed by a ReLU non-linearity followed by max-pooling. Every convolutional layer, had 64 filters, a padding of 1, a stride of 1 and a filter size of 3. 

In the case of MAML, following the convolutional network, we applied a linear layer and a softmax activation function to produce the network's predictions. The size of the last layer is equal to the number of classes per set.

In the case of Prototypical Networks, the networks predictions are the flattened outputs of the fourth convolutional layer, exactly as done in the original paper.





\section{Results}
Tables \ref{table:compare-mini-imagenet} and \ref{table:compare-omniglot} present comparative results between a variety of baseline and SOTA methods for unsupervised few shot learning. Our proposed method outperforms all other methods across all Omniglot tasks, and it remains competitive in the Mini-Imagenet 5-way 1-shot task. In the 5-way 5-shot Mini-Imagenet our method is outperformed by the CACTUs methods. Furthermore, to investigate the effect of the data augmentation strategies on the generalization performance of our method we conducted ablation studies. Tables \ref{table:umaml-ablation-omniglot}, \ref{table:uproto-ablation-omniglot}, \ref{table:umaml-ablation-mini-imagenet} and \ref{table:uproto-ablation-mini-imagenet} showcase the results of the ablation studies. Our ablation studies revealed that for the Omniglot task, the most useful data augmentation methods are pixel-dropout and warping. Warping produces state of the art results in Omniglot, we hypothesize that this is due to the augmented samples matching very closely natural variations of the source images. In the Mini-Imagenet tasks we observed that, surprisingly, crops and flips outperformed all other methods. Much of the success of our Omniglot results, came from using augmentations that actually generate a target set that is meaningful yet initially difficult for the underlying meta-learning method to perform on. In Mini-Imagenet, we observed that all the data augmentation strategies attempted were overfitting after just 5 epochs. This indicates that more work is required to find a good augmentation strategy for the Mini-Imagenet tasks.

\section{Conclusion}
Learning robust representations using label-free data is crucial task for any truly intelligent agent. In this paper, we propose a methodology that allows a machine to create its own tasks using unlabelled datasets by leveraging randomly clustered data-points and augmentation. Models trained with our method, are able to generalize very well to real-labeled inference tasks. 
We compared our method against a variety of recent unsupervised few-shot methods. We evaluated our method on Omniglot and Mini-Imagenet. Our model outperformed all other competitors in Omniglot, whilst remaining competitive in the 5-way 1-shot Mini-ImageNet task. However, in the 5-way 5-shot task, our method was outperformed by CACTUs. Finally, a very noteworthy observation is that the performance of our method is heavily dependant on the data augmentation strategies employed. Hence, finding the most optimal one is key. Automating the search for such a data augmentation function could produce fully unsupervised systems with stronger performance.

\section{Acknowledgements}
We would like to thank Harri Edwards for supporting this project through useful discussions and suggestions. Furthermore, special thanks to Elliot Crowley and Joseph Mellor for their writing-related suggestions and corrections. Finally, we'd like to thank Luke Nicholas Darlow for the very useful discussions on random labels and their intriguing properties. This work was supported in by the EPSRC Centre for Doctoral Training in Data Science, funded by the UK Engineering and Physical
Sciences Research Council and the University of Edinburgh as well as by the European Unions Horizon 2020 research and innovation programme under grant agreement No 732204 (Bonseyes) and by the Swiss State Secretariat for Education Research and Innovation (SERI) under contract number 16.0159. The opinions expressed and arguments employed herein do not necessarily reflect the official views of these funding bodies.

\bibliography{deeplearning,metalearning,datasets,general}
\bibliographystyle{icml2019}





\end{document}